\documentclass{article}
\usepackage{spconf,amsmath,graphicx}

\usepackage{enumitem}
\usepackage{amssymb}
\usepackage{multirow}
\usepackage{tabularx}
\usepackage{comment}
\setlist{nosep, leftmargin=14pt}

\usepackage{mwe} 

\title{Structure-augmented standard plane detection with temporal aggregation in blind-sweep fetal ultrasound} 
\name{Keli Niu$^{\dagger}$ \quad He Zhao$^{\ddagger}$ \quad Qianhui Men$^{\dagger}$}
\address{University of Bristol}
\address{$^{\dagger}$ University of Bristol, $^{\ddagger}$ University of Liverpool}

\begin{document}
%
\maketitle

\begin{abstract}
    In low-resource settings, blind-sweep ultrasound provides a practical and accessible method for identifying fetal growth restriction. However, unlike freehand ultrasound which is subjectively controlled, detection of biometry plane in blind-sweep ultrasound is more challenging due to the uncontrolled fetal structure to be observed and the variaties of oblique planes in the scan. In this work, we propose a structure-augmented system to detect fetal abdomen plane, where the abdominal structure is highlighted using a segmentation prior. 
    Since standard planes are emerging gradually, the decision boundary of the keyframes is unstable to predict. We thus aggregated the structure-augmented planes with a temporal sliding window to help stabilise keyframe localisation.
    Extensive results indicate that the structure-augmented temporal sliding strategy significantly improves and stabilises the detection of anatomically meaningful planes, which enables more reliable biometric measurements in blind-sweep ultrasound.

\end{abstract}

\section{Introduction}

Fetal growth restriction (FGR) is a leading cause of perinatal morbidity and mortality, affecting about one in ten pregnancies worldwide \cite{bernstein2000morbidity-2}. For the early detection of FGR, it heavily relies on accurately identifying standard planes in obstetric ultrasound (US) to ensure precise fetal biometric measurements~\cite{salomon2022isuog-15}. 
However, due to a shortage of skilled sonographers in low-resource settings \cite{sappia2025acouslic-3}, blind-sweep protocols \cite{self2022developing} which scan along predefined trajectories have become a practical approach for estimating gestational age and detecting fetal abnormalities. These protocols call for accurate and reliable detection of biometry planes for measurement.
Existing standard plane detection approaches \cite{baumgartner2017sononet,pu2021automatic,zeng2023tuspm,he2024fetal} mainly focus on multi-task learning, where the semantic understanding of the anatomical structures within the plane is learned simultaneously.  
One of the recent approaches~\cite{guo2024mmsummary} adopted biomedical imaging foundation models like BiomedCLIP~\cite{zhang2023biomedclip-13} to learn the fetal image semantics, which accurately retrieves standard planes in less challenging freehand US scanning videos. However, performance of these models downgrades on blind-sweep videos, where low-quality, non-standard planes or partially visible standard planes are common.
This is because keyframe detection in blind-sweep US requires not only semantic understanding but also the ability to recognise and localise fine-grained anatomical structures, which is typically conducted by experienced sonographers in freehand US.

\begin{figure*}[htb]
  \centering
  \centerline{\includegraphics[width=0.85\linewidth]{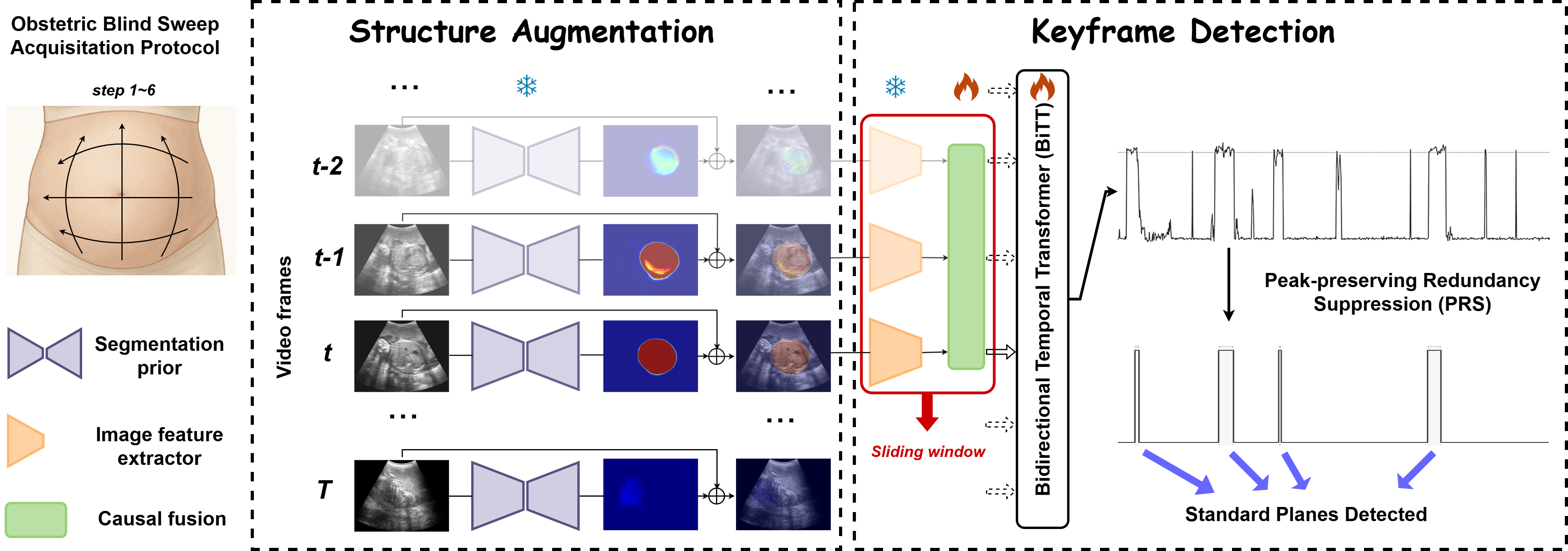}}
%
\caption{Overview of structure-augmented standard plane detector with temporal localisation.}
\label{fig:pipeline}
\end{figure*}

In this work, we propose a structure-augmented keyframe detector to solve the challenges of standard plane detection in blind-sweep scan. The main architecture first consists of a structural prior to detect and preserve the visual saliency from the video frames. 
The structural prior augments each video frame with anatomical saliency derived from a separately trained nnU-Net–based~\cite{isensee2021nnu} segmentation backbone. 
Given that standard planes in blind-sweep US often appear gradually, directly detection of temporal boundary near the standard planes usually provides unstable results. 
We therefore adopt a multi-frame local aggregation strategy with a sliding window across the video frames, where the current frame is integrated with its preceding frames to enhance the representation of the current anatomical structures. This aggregation step helps stabilise the keyframe detection and localisation. This is because, for example, if the features in preceding frames are anatomically irrelevant, the current frame is unlikely to mark the start of a standard-plane segment. 
A bidirectional temporal transformer (BiTT) is then used as the keyframe detector for global contextual standard plane detection from the spatial-temporal enhanced structural features.
Evaluated on real-world obstetric blind sweep videos, our keyframe detector yields consistent gains over representative segmentation and semantic modelling baselines. An overview of our structure-augmented temporal aggregation framework is shown in Fig.~\ref {fig:pipeline}.

\section{Methodology}
\label{sec:method}

We cast keyframe localisation in blind–sweep US as a \emph{binary} sequential labelling task. 
Given a 2D B\mbox{-}mode video $X=\{I_t\}_{t=1}^T$ with $T$ frames, we predict per\mbox{-}frame posterior $p_t=P(y_t=1 \mid I_{1:T})$, where $y_t=1$ indicates a standard abdominal plane.
Our pipeline (Fig.~\ref{fig:pipeline}) for standard plane detection has two main stages: 
\textbf{(i) Structure augmentation:} a pretrained segmentation model applied on each US frame to identify the presence of abdominal structures and highlight them when present; 
\textbf{(ii) Keyframe detection:} a sliding window with \emph{causal} fusion moving forward across the temporal frames to reduce motion noise in US frames and enhance the spatial evidence obtained in the previous step,
and a bidirectional temporal transformer to estimate the global contextual probability of each frame belonging to a standard plane. During inference, the probability output is refined by a peak-preserving module, which selects the frames with the strongest evidence while removing redundancy within the detected segments.

\subsection{Structure-Augmented Segmentation Prior}
Detecting abdominal planes in obstetric US relies on recognising anatomical landmarks such as stomach bubble and umbilical vein~\cite{salomon2022isuog-15}. Blind-sweep sequence, however, contains extensive background noise and distracting non-target structures, which obscure the cues for measuring the targets. 

We therefore propose to augment clinical meaningful anatomies in the blind-sweep frames using a structural segmentation prior.
Specifically, a segmentation model with nnU-Net~\cite{isensee2021nnu} backbone is trained on the training split to predict a frame-wise binary abdominal mask $M_t$ ($1$ inside the abdomen and $0$ elsewhere). $M_t$ is then converted into a simple structural prior by amplifying pixel intensities \emph{only} inside the segmented region before feature extraction. The augmented image frame is represented as
$I_t^{\ast} \;=\; I_t \odot \big(1 + \alpha\, M_t\big)$,
where $\alpha$ controls the strength.
The mask acts purely as prior structural information and does not alter labels or the learning objective. 
The frozen prior avoids the need for dense frame annotations and allows real-time processing during inference.

\subsection{Keyframe Detection}

\subsubsection{Multi-depth Spatial Encoding}
Leveraging the strong capability of image foundation models in understanding biomedical visual semantics, we use a pretrained BiomedCLIP Vision Transformer (ViT\mbox{-}B)~\cite{zhang2023biomedclip-13} as the image extractor backbone, inspired by ~\cite{guo2024mmsummary}. 
Given that US standard planes always contain complementary visual cues, i.e., from low-level boundary details to high-level semantics with anatomical significance, we learn these complementary image features by encoding and fusing the representations 
from the different depths of ViT outputs. This design follows the principle of Feature Pyramid Networks (FPN)~\cite{lin2017feature} but adapting multi-scale to multi-depth representation fusion, where shallow layers focus on edges and landmarks, and deeper layers capture global contours and contextual consistency.
In each modelled layer, the extracted feature passes through a Squeeze-and-Excitation (SE) block~\cite{hu2018squeeze} to reweight feature channels beyond the pixel-level emphasis of the prior.

Specifically, the image feature extracted from the structure-augmented frame \(I^{\ast}\) is represented as \(\{ f^{(l)}\}_{l=1}^{L}\), where $l$ is the layer depth of ViT. 
SE block further rescales the feature as \(\tilde f^{(l)}=s\odot f_t^{(l)}\) with the learned channel weights \(s\in(0,1)^{D_l}\), and $D_l$ is the channel dimension at depth $l$. 
The multi-layer representations are fused in a recursive manner, i.e., \(r^{(1)}=\tilde f^{(1)}\), \( r^{(l)}=\mathrm{MLP}([\mathrm{Proj}(r^{(l-1)}),\,\tilde f^{(l)}])\) for $2\leq l\leq L$, where \([\cdot,\cdot]\) denotes concatenation, \(\mathrm{Proj}(\cdot)\) is a linear projection, and \(\mathrm{MLP}\) is a two-layer perceptron. We learn and merge the feature at depth $l$ from deep to shallow at layers 5, 7, 9, and 11. 

\subsubsection{Local and Global Temporal Conditioning}
\textbf{Local sliding window.}
Given that US standard planes appear progressively, single-frame decisions near the boundaries of standard-plane segments are often unstable. Prior work stabilises detections 
by reading from an external space–time memory \cite{oh2019video}. However, these approaches introduce extra signals and latency, which are ill-suited for blind-sweep scenarios that require lightweight architectures. Here, we adopt a parameter-saving approach by aggregating temporal history with a short sliding window, i.e., $f_t=\sum_{i=0}^{k-1} w_{i}f_{t-i}, w_i \ge 0$ and $\sum_{i=0}^{k-1} w_i=1$. The frame weights \(\{w_i\}\) are initialised with a monotonically decaying profile such that the current frame dominates. This is implemented as a depthwise 1D convolution along time, with the window size $k$ as \textit{kernel}
and \textit{left padding} exclusively, which preserves the chronological order of video sequences and maintains causality.

\textbf{Global temporal contexts.}
We adopt a bi-directional temporal Transformer (BiTT) to capture global dependencies along the blind-sweep scan and to localise the moments when standard planes appear.
Unlike the vanilla Transformer that fuses past and future information in a single stream, our approach enforces directionality with two unidirectional streams. Together with bidirectional beam search, this yields improved sequence decisions~\cite{chen2020transformerbidirectionaldecoderspeech}. Explicitly,  
a forward transformer encodes \(X\) to produce temporal feature sequence \(H^{\mathrm f}\) and a second transformer encodes \(\mathrm{flip}(X)\) with the output flipped back as \(H^{\mathrm b}\). We concatenate \(H=[H^{\mathrm f};\,H^{\mathrm b}]\) and apply a classification head to output per-frame logits. 
Unlike masked autoencoders (MAE), our temporal transformer is trained directly for frame-wise detection on image features. The bidirectional streams provide long-range context from both past and future frames, reducing boundary ambiguity.

\subsubsection{Peak-Preserving Redundancy Suppression (PRS)}
To mitigate temporal jitter and short-lived spurious peaks while preserving true peaks, we propose PRS, a post-hoc, peak-preserving smoothing method that consolidates detections into compact segments. 
\emph{(1) Smoothing:} we apply Savitzky--Golay (SG) 
and an exponential moving average (EMA) 
to the posterior sequence to obtain \(p_{\mathrm{sg}}\) and \(p_{\mathrm{ema}}\). A convex weight \(\lambda\in(0,1)\) balances short-term detail of SG against longer-term stability of EMA, and a small downward clamp \(\delta>0\) prevents shaving strong peaks.
Let \(z_t\) be the output logit from BiTT, and let \(p(t)=\sigma(z_t)\in(0,1)\) denotes its probability at $t$. The smoothing step is denoted as:
 
\begin{equation}
p_{\mathrm{smooth}}(t)=\max\!\big\{\lambda\,p_{\mathrm{sg}}(t)+(1-\lambda)\,p_{\mathrm{ema}}(t),\; p(t)-\delta\big\}.
\end{equation}
\emph{(2) Grouping:} threshold \(p_{\mathrm{smooth}}\) at \(\theta\), i.e., \(p_{\mathrm{\theta}}=\mathbb{I}(p_{\mathrm{smooth}}\ge \theta)\), remove very short segments and merge those separated by small gaps to form dwell-like segments. 
\emph{(3) Peak preservation:} detect local maxima on \(p_{\mathrm{\theta}}\), expand each within a small window, and retain a candidate only if intermediate frames meet a local cosine-similarity check to the peak centre. 
Frames inside the final segments are labelled \(\hat y_t{=}1\) and others \(0\). If no segment is found, we promote \(\arg\max_t p(t)\).

\subsection{Training Objective}
\label{sec:myloss}
We train the detector as frame-wise \emph{binary} classification using a weighted Binary Cross Entropy (BCE) loss $\mathcal{L}_{\mathrm{BCE}}$ on the logit $z_t$
as the primary objective. We also include two supplementary losses to enlarge margins and improve calibration without changing label semantics, i.e., a pairwise margin-based contrastive loss and a soft-target regulariser.

\textit{Pairwise margin-based contrastive loss $\mathcal{L}_{\mathrm{con}}$}. 
Let $H=\{\mathbf{h}_t\}_{t=1}^T$ be $\ell_2$-normalised embeddings and 
$\mathcal{P}=\{(i,j)\mid 1\le i<j\le T\}$. 
Define $d_{ij}=\lVert \mathbf{h}_i-\mathbf{h}_j\rVert_2$, $y_i$ as the binary label at frame $i$ and margin $m>0$, 
we have
\begin{equation}
\mathcal{L}_{\mathrm{con}}
=\frac{1}{|\mathcal{P}|}\sum_{(i,j)\in\mathcal{P}}
\Big(\mathbb{I}[y_i{=}y_j]\; d_{ij}^{2}
+\mathbb{I}[y_i{\neq}y_j]\; [\,m-d_{ij}\,]_+^{2}\Big).
\end{equation}

\textit{Soft-target regularisation $\mathcal{L}_{\mathrm{gs}}$.}
We apply a two\mbox{-}class Gumbel\mbox{--}Softmax relaxation~\cite{jang2016categorical} to the same logit \(z_t\) with temperature \(\tau>0\).
With i.i.d.\ noise \(g_0,g_1\!\sim\!\mathrm{Gumbel}(0,1)\), the \emph{training\mbox{-}only} soft foreground probability is
\begin{equation}
\tilde p_t=\frac{\exp\!\big((z_t+g_{1})/\tau\big)}{\exp\!\big(g_{0}/\tau\big)+\exp\!\big((z_t+g_{1})/\tau\big)}.
\end{equation}
\(\tilde p_t\) is then matched to \(y_t\) via BCE.

To stabilise the multi-depth spatial encoding, an auxiliary BCE $\mathcal{L}_{\mathrm{aux}}$ is used for deep supervision of each fusion head. Moreover, an optional short-range smoothness penalty on softened probabilities mitigates frame-to-frame jitter $\mathcal{L}_{\mathrm{tmp}}$. The final objective is a weighted combination of the five training losses.

\section{Experiments}

\subsection{Dataset and Implementation Details.}
We evaluate on the open-sourced real-world obstetric US dataset~\cite{sappia_2024_12697994-data-15}, which contains 300 blind-sweep cases, each with six sweeps and 840 frames with a resolution of 744$\times$562. Frame label \textit{BG} denotes background, \textit{Key} and \textit{Sub} are grouped as foreground label in our experiment.
We split the total scans into 210/45/45 for train/validation/test set (case-level split). 
In the segmentation prior, the abdominal pixels are highlighted with $\alpha{=}0.15$. 
The size of the local sliding window $k$ is 4.
Adam is used as training optimiser with a learning rate of $10^{-4}$. The temperature $\tau$ for Gumbel--Softmax regularisation is annealed from 1.5 to 0.5 with a decay factor of 0.98 and $m$ in the contrastive loss is 1. 
We train for 150 epochs with batch size 1 and early stopping is applied. 
For PRS, min segment length is 3 frames, max gap is 2, convex weight $\lambda$ is 0.6, downward clamp $\delta$ is 0.05, and threshold $\theta$ is 0.9.

\begin{table}[htb]
\centering

\caption{Quantitative result comparisons for keyframe detection in blind-sweep US. SA: structure-augmented segmentation prior. MS: multi-depth spatial encoding. MT: local and global temporal conditioning. Best per column in \textbf{bold}.}
\label{tab:main_binary}
\resizebox{1.0\columnwidth}{!}{
\begin{tabular}{lccccc}
\hline
\multirow{2}{*}{Method} & P & R & F1 & Abs. time & Keyf. num. \\
& (\%) & (\%)& (\%)& err. (↓)& err. (↓)\\
\hline
BiomedCLIP & \multirow{2}{*}{37.83} & \multirow{2}{*}{51.12} & \multirow{2}{*}{43.48} & \multirow{2}{*}{46.47} & \multirow{2}{*}{14.00} \\
+ViT-B & & & & & \\
nnU\mbox{-}Net & 52.65 & 63.58 & 57.60 & 77.78 & 10.07 \\\hline
\textit{MS} & 40.21 & 45.47 & 42.68 & 60.57 & 11.80 \\
\textit{MT} & 46.55 & 41.00 & 43.60 & 49.29 & 11.29 \\
\textit{MS+MT} & 43.73 & 47.92 & 45.73 & 52.59 & 11.2 \\
\textit{SA+MS+MT} & 60.14 & \textbf{72.63} & 65.80 & 16.82 & 10.91 \\
\textit{SA+MS+MT+PRS} & 61.11 & \textbf{72.63} & 66.37 & 16.45 & 10.82 \\
\textit{SA+MS+MT+PRS+$\mathcal{L}$} & \textbf{64.20} & 70.07 & \textbf{67.01} & \textbf{15.33} & \textbf{10.04} \\
\hline
\end{tabular}}
\end{table}

\subsection{Quantitative Evaluation}
\textbf{Metrics.} We report \emph{Precision (P)}, \emph{Recall (R)}, and \emph{F1} on frame-wise labels. Timing fidelity is evaluated by: 
(i) \emph{absolute time error}, which measures temporal misalignment as the mean of $|t_{\text{pred}}-t_{\text{gt}}|$ between the predicted and GT keyframes, which is sensitive to isolated false peaks; and (ii) \emph{keyframe number error}, which is
the discrepancy of lengths between the predicted and GT segments.

Table~\ref{tab:main_binary} compares nnU\mbox{-}Net, BiomedCLIP+ViT\mbox{-}B, and our variants. 
\textbf{Baseline comparisons.} The complete model of our approach (last row) achieves an F1 score of 67.01\%. The substantially higher precision and recall rates compared to the two baselines indicate that our model effectively filters out non-standard-plane frames while being less likely to miss standard planes, which are crucial for accurate downstream biometric measurements in blind-sweep US. The lower time and keyframe errors further demonstrate the precise localisation of our method in identifying standard plane segments.
\textbf{Ablations.} In general, the performance of our methods across the five metrics consistently improves with the structural and temporal modules. 
\emph{MS} alone results in unstable temporal localisation, indicating that the model lacks boundary cues without temporal context. \emph{MT} alone achieves better absolute\mbox{-}time alignment, but the lower recall rate indicates that insufficient spatial discrimination allows some structurally ambiguous frames to pass. 
\emph{MS+MT} balances precision and recall for a boost in F1, 
and adding the structural prior (\emph{SA+MS+MT}) greatly improves the score 
by foreseeing abdominal landmarks, which also resolves boundary ambiguity with a much lower time error obtained. 
Absolute-time error drops sharply because it penalises temporally distant false peaks (e.g., isolated off-plane activations), whereas keyframe-number error mainly reflects segment length mismatch and is less sensitive to such outliers.
With \emph{PRS}, isolated spikes are removed and short gaps are bridged, which further refines the temporal alignment. Finally, the system with the full objective 
(\emph{+$\mathcal{L}$}) further increases precision, yielding the best F1 and the lowest timing error.

\subsection{Qualitative Evaluation}
\begin{figure}[t]
  \centering
  \includegraphics[width=8.5cm]{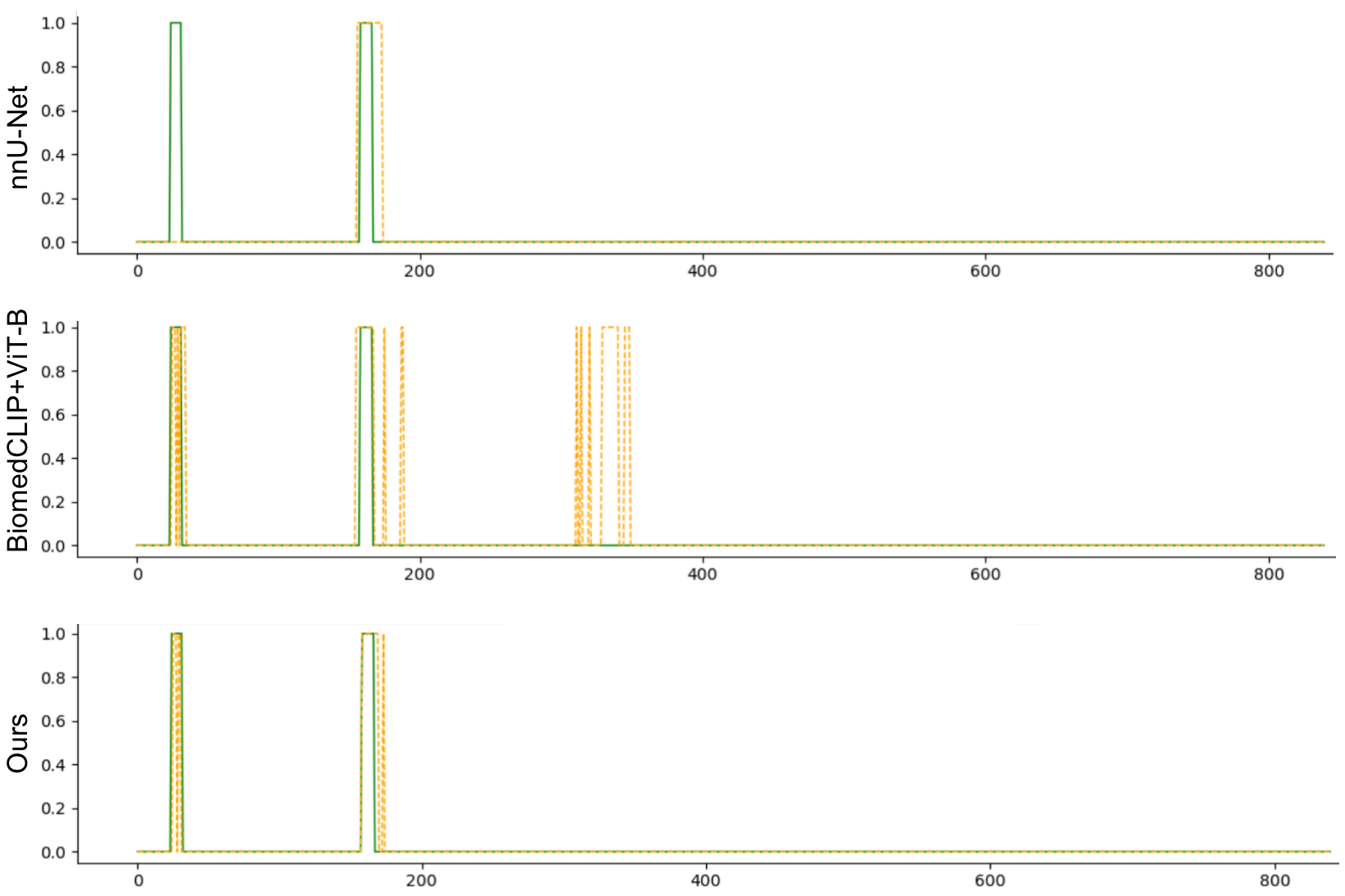}
  \caption{Sequential detection results.
  \textit{Green}: ground truth (GT) standard plane segments; \textit{Orange dashed}: predictions. 
  }
  \label{fig:qual_traces}
\end{figure}

\noindent\textbf{Sequential detection.}
Fig.~\ref{fig:qual_traces} compares per\mbox{-}frame labelling results on an example blind-sweep scanning sequence. 
For the two real standard plane segments, nnU-Net clearly missed one, whereas BiomedCLIP+ViT\mbox{-}B produced multiple off\mbox{-}plane spikes. Our prediction is closest to the GT, with strong boundary fidelity and robust noise rejection.

\begin{figure}[t]
  \centering
  \includegraphics[width=8.5cm]{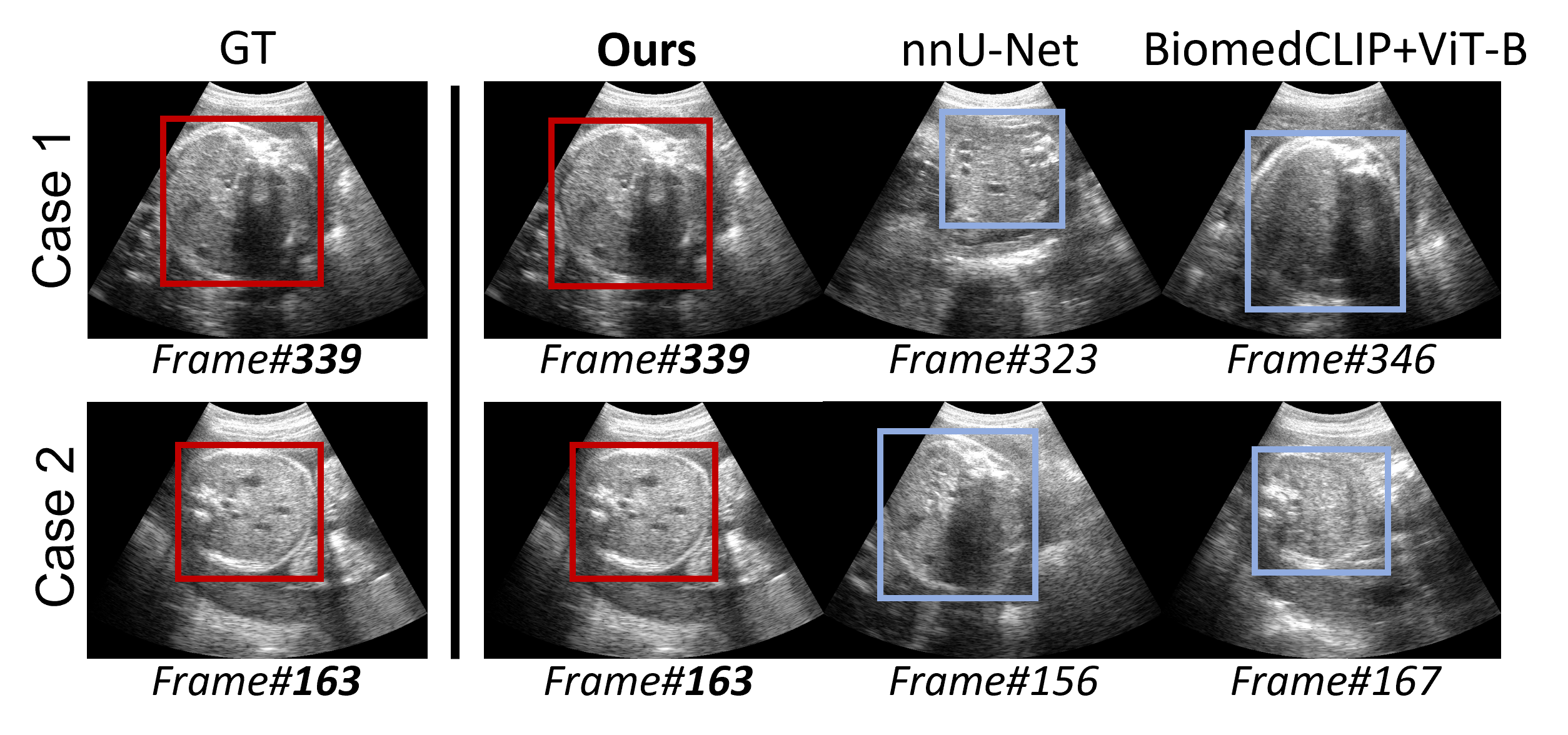}
  \caption{The detected keyframes of two example scans with the abdominal anatomy highlighted by bbox.}
  \label{fig:qual_frames}
\end{figure}

\noindent\textbf{Keyframe visualisation.}
Fig.~\ref{fig:qual_frames} compares the top-scored frames of each model from the output probability, and their corresponding frame numbers in the scan.
Our selections consist with the GT (i.e., frame \#339 and \#163); nnU\mbox{-}Net and BiomedCLIP+ViT\mbox{-}B tend to match temporally misaligned look\mbox{-}alike frames, and their anatomical structures are visually inconsistent to the GT standard planes.

\section{Conclusion}
We propose a structure\mbox{-}augmented keyframe detector for fetal standard abdominal planes in the blind\mbox{-}sweep obstetric US. The abdominal structures are highlighted in the image frames using a segmentation prior, which allows the model to focus on anatomically meaningful regions during the detection and greatly reducing the false positive rates. Since the gradual emergence of the standard plane usually leads to uncertainty in the temporal detection, we introduce a local sliding window to stabilise the standard-plane decision boundaries, together with a posterior smoothing strategy to reduce temporal jitter in the predictions.
Quantitative and qualitative results on a real-world blind-sweep dataset indicate accurate boundary localisation and fewer off-plane activations, supporting more reliable downstream fetal biometry. 
\section{Compliance with ethical standards}
\label{sec:ethics}
This study was conducted retrospectively using ethically acquired publicly available human subject data. The authors have no interests to disclose.

\bibliographystyle{IEEEbib}
\bibliography{refs}

\end{document}